\documentclass[aps,prl,amsmath,amssymb,twocolumn,superscriptaddress]{revtex4-2} 

\usepackage{graphicx}
\usepackage{dcolumn}
\usepackage{bm}
\usepackage{lipsum} 

\begin{document}

\title{Large Language Models as Quasi-crystals: Coherence Without Repetition in Generative Text}
\author{Jos\'e Manuel Guevara-Vela}
\affiliation{School of Engineering and Physical Sciences, Heriot-Watt University, Edinburgh EH14 4AS, Scotland, U.K. \\ \texttt{email: j.guevara-vela@hw.ac.uk}}
\email{j.guevara-vela@hw.ac.uk}

\date{\today}

\begin{abstract}

This essay proposes an interpretive analogy between large language
models (LLMs) and quasicrystals, systems that exhibit global coherence without
periodic repetition, generated through local constraints. While LLMs are
typically evaluated in terms of predictive accuracy, factuality, or alignment,
this structural perspective suggests that one of their most characteristic
behaviors is the production of internally resonant linguistic patterns. Drawing
on the history of quasicrystals, which forced a redefinition of structural
order in physical systems, the analogy highlights an alternative mode of
coherence in generative language: constraint-based organization without
repetition or symbolic intent. Rather than viewing LLMs as imperfect agents or
stochastic approximators, we suggest understanding them as generators of
quasi-structured outputs. This framing complements existing evaluation
paradigms by foregrounding formal coherence and pattern as interpretable
features of model behavior. While the analogy has limits, it offers a
conceptual tool for exploring how coherence might arise and be assessed in
systems where meaning is emergent, partial, or inaccessible. In support of this
perspective, we draw on philosophy of science and language, including
model-based accounts of scientific representation, structural realism, and
inferentialist views of meaning. We further propose the notion of structural
evaluation: a mode of assessment that examines how well outputs propagate
constraint, variation, and order across spans of generated text. This essay
aims to reframe the current discussion around large language models, not by
rejecting existing methods, but by suggesting an additional axis of
interpretation grounded in structure rather than semantics.

\end{abstract}

\maketitle

\section{Introduction}

Large language models (LLMs) are statistical systems trained on massive corpora
of human text to generate coherent language outputs. Developed through deep
learning architectures such as the Transformer,~\cite{vaswani2017attention}
these models operate by predicting the most probable next token in a sequence
based on prior context. A token is a basic unit of text (typically a word,
subword, or punctuation mark) used to segment language for computational
modeling. Each new token is selected in relation to all those that came before,
allowing the model to generate fluent sequences without predefined grammar or
semantic rules. Since the release of GPT-2 in 2019~\cite{radford2019language}
and especially GPT-3 in 2020,~\cite{brown2020language} which helped inaugurate
the class of foundation models,~\cite{bommasani2021opportunities} LLMs have
shown an ability to produce language that is locally fluent and contextually
plausible across a wide range of domains. These outputs are not the result of
symbolic reasoning or understanding, but of pattern recognition applied at
large scale. The resulting text often exhibits internal consistency, thematic
alignment, and the appearance of intention, even when none is explicitly
encoded.~\cite{bender2021dangers}

Quasicrystals are a class of solid materials that exhibit long-range order
without periodic repetition. First identified in 1982 by Dan
Shechtman,~\cite{shechtman1984metallic} their discovery challenged the
prevailing assumption that atomic arrangements in crystals must repeat
regularly in space. Quasicrystals display symmetry, such as fivefold rotations,
that are forbidden in classical crystallography, yet their structure is neither
random nor disordered. Instead, they possess deterministic but aperiodic
arrangements, as first modeled by Levine and Steinhardt using tiling patterns
such as those of Roger Penrose.~\cite{levine1984quasicrystals} The recognition
of quasicrystals as a legitimate form of structural order expanded the
conceptual boundaries of solid-state chemistry and materials science, opening
new directions in the study of symmetry, entropy, and
self-assembly.~\cite{senechal1996quasicrystals}

We propose that the structure of LLM-generated text exhibits properties
analogous to those of quasicrystals. The analogy applies specifically to the
output: sequences of language produced through token-level prediction, not the
internal architecture of the models themselves. Like quasicrystalline matter,
this output displays local coherence without global repetition. It is neither
disordered nor strictly periodic, but unfolds according to statistical
constraints that produce a recognizably structured, yet non-repeating,
surface.~\cite{blier2018description} Just as quasicrystals forced a
redefinition of what constitutes order in physical systems, LLMs challenge
inherited assumptions about coherence in language. Other systems, such as
cellular automata, also generate complex behavior from simple local rules and
have been widely studied as models of emergent structure.~\cite{wolfram2002new}
However, quasicrystals are distinctive in that they introduced a formally
recognized category of long-range aperiodic order in the physical sciences.
This makes them especially apt as analogues for linguistic outputs that exhibit
coherence without explicit rules or symbolic intent. In both cases, structured
form emerges not from central planning, but from distributed interactions that
propagate constraint across a generative space.~\cite{wei2022emergent}

This analogy offers more than surface resemblance, it can clarify why LLM
outputs exhibit a surprising degree of coherence without explicit planning, and
why they resist both traditional linguistic analysis and formal logical
modeling. While terms like ``sense'' and ``coherence'' remain philosophically
open, we use them here in a structural sense: to describe the appearance of
internal consistency, recurrence, and thematic alignment across spans of
generated text. This perspective offers a middle ground between dismissive
accounts that see LLMs as stochastic parrots~\cite{bender2021dangers} and
overly anthropomorphic ones that frame them as emergent
minds.~\cite{wei2022emergent} By emphasizing their capacity to generate
constraint-based linguistic forms, we gain a language for describing the
generative space LLMs inhabit, one that accounts for novelty without invoking
consciousness. This shift may support more precise discussions of authorship,
creativity, and trust in machine-generated
language,~\cite{bommasani2021opportunities} as well as inform future model
design across domains that demand both structure and novelty.

The analogy we draw is not intended as an exact isomorphism between physical
and linguistic systems, but as a model-based framework in the sense described
by Hesse~\cite{hesse1966models} and Suárez~\cite{suarez2004representation}: a
heuristic device that preserves relevant structure and supports new modes of
inquiry. In this case, the quasicrystal model provides a way to conceptualize
LLM outputs as emergent structures governed by local constraint rather than
global intent. As with many productive scientific analogies, the value lies not
in literal correspondence but in enabling reframing and making visible aspects
of coherence and generation that standard evaluations might overlook.

The remainder of this essay unfolds in three parts. We first summarize the
relevant properties of quasicrystals and LLMs, with particular attention to how
structure arises in each system from local constraints. We then develop the
analogy between them, highlighting how coherence and form emerge without
repetition or centralized control. Building on this, we outline practical
implications for evaluating and training language models, suggesting that their
behavior may be better understood in structural rather than semantic terms.
Finally, we conclude with a reflection on how this analogy reframes questions
of coherence, meaning, and form in generative systems more broadly.

\section{Order without periodicity}

Quasicrystals display a form of structural order that emerges without
translational periodicity. Unlike conventional crystals, which repeat unit
cells in regular three-dimensional arrays, quasicrystals arrange atoms
according to deterministic rules that never produce a repeated pattern. This
leads to symmetries considered forbidden in classical crystallography, such as
fivefold rotations.~\cite{shechtman1984metallic} Their diffraction patterns contain
sharp Bragg peaks, confirming long-range order, but they cannot be indexed
using the lattice parameters of periodic crystals.~\cite{levine1984quasicrystals}

To describe these materials, scientists have relied on mathematical tiling
schemes, such as Penrose tiling in two dimensions and Ammann tilings in
three.~\cite{senechal1996quasicrystals} These models break space into basic
geometric shapes—tiles—that must fit together according to specific local
rules. Unlike in periodic crystals, these tilings never repeat exactly, yet
they still generate long-range order. The local constraints propagate outward
to form globally structured, non-periodic arrangements. The resulting patterns
maintain coherence across large distances without regular repetition, revealing
a form of organization that sits between randomness and strict regularity. The
recognition of this category required a broadening of how structural order is
defined, both in physical systems and in the conceptual language of
symmetry.~\cite{senechal1996quasicrystals,levine1984quasicrystals}

Large language models generate text by predicting the most likely next token in
a sequence, given a preceding context. This process is not governed by rules of
syntax or meaning in the traditional sense, but by probabilistic correlations
learned from large datasets of human language. Each output is the result of
locally conditioned prediction: the model selects a token based on the
statistical structure of the tokens before it, with no explicit awareness of
global structure or intent (though in practice, the attention mechanism grants
access to the full preceding context). Yet, the generated language often
appears coherent across multiple sentences or even paragraphs, suggesting a
kind of emergent organization.~\cite{blier2018description}

This emergent structure arises from the scale and architecture of modern LLMs.
With billions of parameters and attention mechanisms that capture relationships
between distant tokens,~\cite{vaswani2017attention} these models encode a
diffuse statistical map of language use.~\cite{bommasani2021opportunities}
Coherence is not programmed, but surfaced. It is a byproduct of local decisions
cascading under constraint. As with tiling schemes in quasicrystals, small,
local rules can give rise to patterns that feel organized, readable, and
meaningful over extended spans, even in the absence of explicit design. The
result is a form of language that is neither scripted nor random, but ordered
in ways that resist simple repetition or deterministic
unfolding.

Despite operating in different domains, material and linguistic, quasicrystals
and large language models both generate structured patterns through systems of
local constraint.~\cite{levine1984quasicrystals} In
quasicrystals, tiling rules enforce spatial relationships that propagate
non-repeated, ordered patterns. In LLMs, probabilistic models of token
relationships produce text that unfolds in coherent but non-deterministic ways.
Neither system relies on a global plan or repeated unit, yet both give rise to
forms that are internally consistent and often perceived as intentional or
meaningful. This resemblance is more than superficial: it suggests a shared
structural logic that may help explain the growing sense that LLM-generated
language ``makes sense'' even when it is not grounded in symbolic
understanding.

This analogy does not imply equivalence between material and linguistic order,
but instead highlights a zone of structural similarity where pattern emerges
without repetition and coherence does not require blueprinting. Considering LLM
outputs as linguistic analogues to quasicrystalline structure offers a new
framework for interpreting their generative behavior not as noise, nor mimicry,
but as a distinct form of constrained emergence. This framing may inform how we
evaluate, design, and trust generative systems, not by replacing traditional
metrics of logic or narrative, but by supplementing them with attention to
structural coherence, propagation, and form.

\section{Toward Structural Evaluation}

Recent claims about LLMs' capacity for reasoning, planning, or
logic~\cite{lewkowycz2022solving, wei2022emergent} might appear to challenge
this framing. However, we suggest that such behaviors are better understood as
high-level manifestations of statistical coherence, rather than evidence of
symbolic reasoning in the traditional sense. As others have argued, LLMs
generate plausible approximations of reasoning by operating over patterns
learned from text, not by manipulating abstract concepts or representations
with intent.~\cite{bender2021dangers, marcus2022deep} Our analogy focuses on
this structural dimension: the model’s ability to produce outputs that exhibit
internal consistency and pattern without invoking an explicit logic or plan.

This structural perspective also exposes a limitation in how large language
models are currently evaluated. These systems are typically assessed through
functional lenses: accuracy in downstream tasks, consistency with human
preferences, or alignment with external
truths.~\cite{bommasani2021opportunities,openai2024gpt4} While these metrics
are valuable, they often treat coherence as a secondary byproduct of
optimization rather than as a central feature. If we take seriously the analogy
between LLMs and quasicrystals of language, then coherence is not incidental
but fundamental. In this view, an LLM is not merely a predictor of plausible
continuations, but a generator of constrained structure. Its success lies not
only in what it says, but in how its internal form sustains tension, variation,
and order over time.

This reframing also suggests new directions for evaluation. If LLMs are
generative systems whose characteristic behavior is the propagation of internal
structure, then we might assess their outputs in terms of formal properties
rather than semantic correctness. Structural features such as local-to-global
coherence, the recurrence of non-identical motifs, or the statistical rhythm of
variation become relevant indicators. Methods inspired by signal processing,
including entropy mapping, motif extraction, or Fourier-like decomposition,
could surface organizational patterns that token-level metrics fail to detect.
For instance, an entropy heat-map of a 1000-token completion could reveal zones
of high and low unpredictability, indicating shifts in internal structure,
theme, or register. Recurrent but non-identical motifs might suggest narrative
pacing or topical development. Such features would allow for diagnostics that
go beyond correctness to reflect how well a model sustains coherent variation
over time. In this light, structural resonance, the degree to which an output
holds together through internal patterning, might become a meaningful
evaluative target, particularly in domains such as storytelling, dialogue, or
abstraction, where formal coherence is often more valuable than semantic
precision.

Seen through this lens, structural evaluation becomes a complementary mode of
assessment. It does not replace truthfulness, safety, or usefulness as metrics,
but foregrounds a different kind of judgment: whether the output of a model is
internally consistent, non-redundant, and generatively coherent over time. This
approach does not ask whether a model is correct in some external sense, but
whether its outputs maintain formal properties, such as constraint propagation
or emergent patterning, that make them intelligible and interpretable to
humans. In contexts where narrative flow, abstraction, or long-range coherence
are key, this kind of structural fidelity may offer insights that token-level
accuracy cannot. It opens a path for evaluating systems not just by what they
say, but by how they sustain rhythm, pattern, and conceptual balance—qualities
that matter both aesthetically and cognitively, even in the absence of explicit
meaning.

\section{Conclusions}

This essay has proposed a structural analogy between large language models and
quasicrystals, two systems in which coherence emerges from local constraint
without global repetition. Rather than viewing LLMs as imperfect communicators
or stochastic approximators, we have suggested understanding them as generators
of constrained, non-periodic form. This framing contributes to an ongoing shift
in how language and generative systems are conceptualized. In the philosophy of
language, it aligns with inferential and structuralist accounts, where meaning
arises from relational patterns rather than reference
alone.~\cite{sellars1954inferential} In the philosophy of science, it echoes
structural realism, which holds that the epistemic value of scientific theories
often lies in the patterns they preserve, not the entities they
posit.~\cite{ladyman2007every} In AI ethics, it suggests that evaluation itself
is not neutral: shifting our metrics toward structural coherence may shape how
we justify trust, transparency, and interpretability in generative
systems.~\cite{floridi2016dignity,mittelstadt2016ethics,coeckelbergh2020ai}
Just as quasicrystals expanded the scientific definition of order, LLMs may
prompt a broader conception of coherence in language and generation.

This perspective invites us to reconsider what it means for language to make
sense, and how best to engage with systems that produce such sense without
understanding. LLMs do not think in the human sense, but they generate forms
marked by pattern, balance, and constraint, qualities we intuitively associate
with meaning, even though these systems lack intentionality. The question,
then, is not whether such models could give rise to consciousness, but how we
might think about them: how to interpret and design around the distinct kind of
order they instantiate, and how to recognize the opportunities afforded by
their internal structural logic. Like all analogies, this one has boundaries.
It does not imply that LLMs share the physical ontology of quasicrystals, nor
that linguistic structure is reducible to spatial pattern. Rather, it aims to
foreground a neglected mode of evaluation, where structure is coherence without
repetition, and to make space for forms of intelligibility that arise without
intent. The quasicrystal model serves here not as a literal template, but as a
conceptual tool to see coherence where traditional semantics may fall short,
and to frame new questions about design, interpretation, and trust in
generative systems.

\section{References}

%

\end{document}